\documentclass{article}

\PassOptionsToPackage{authoryear,round}{natbib}
    \usepackage[preprint]{neurips_2025}

\usepackage[utf8]{inputenc} % allow utf-8 input
\usepackage[T1]{fontenc}    % use 8-bit T1 fonts
\usepackage{hyperref}       % hyperlinks
\usepackage{url}            % simple URL typesetting
\usepackage{booktabs}       % professional-quality tables
\usepackage{amsfonts}       % blackboard math symbols
\usepackage{nicefrac}       % compact symbols for 1/2, etc.
\usepackage{microtype}      % microtypography
\usepackage{xcolor}         % colors
\hypersetup{
    colorlinks=true,   % <-- 这是关键！设置为 true, 文本变色, 边框消失
    citecolor=blue,    % 将引文颜色设为蓝色
    linkcolor=blue,    % 将内部链接（如章节、公式）也设为蓝色
    urlcolor=blue      % 将URL链接也设为蓝色
}
\usepackage{graphicx}
\usepackage{multirow}
\usepackage{pifont}
\usepackage{bbding}
\usepackage{array}
\usepackage{tikz}
\usepackage{colortbl}   % for table
\usepackage{epigraph}
\usepackage{wrapfig}
\usepackage{makecell}
\usepackage{subcaption}

\usepackage[most]{tcolorbox}
\usepackage{caption}
\usepackage{hyperref}

\usepackage{soul}

\definecolor{code}{HTML}{fefadd}
\definecolor{result}{HTML}{ffe3ec}

\definecolor{code_text}{HTML}{a7adb5}
\definecolor{result_text}{HTML}{a7adb5}
\definecolor{think_text}{HTML}{c9d6f1}
\definecolor{task_text}{HTML}{a7c7e7}

\usetikzlibrary{tikzmark}
\newcommand{\textboxStyle}[2]{%
    \tikzmarknode[draw=#1,thick,fill=white,rounded corners=2pt,inner sep=2pt]{boxnode}{#2}%
}

\newcommand{\resultStart}{\textboxStyle{result_text}{\textbf{\textcolor{result_text}{<execution\_results>}}}}
\newcommand{\resultEnd}{\textboxStyle{result_text}{\textbf{\textcolor{result_text}{</execution\_results>}}}}

\newcommand{\codeStart}{\textboxStyle{code_text}{\textbf{\textcolor{code_text}{<code>}}}}
\newcommand{\codeEnd}{\textboxStyle{code_text}{\textbf{\textcolor{code_text}{</code>}}}}

\newcommand{\taskStart}{\textboxStyle{task_text}{\textbf{\textcolor{task_text}{<task>}}}}
\newcommand{\taskEnd}{\textboxStyle{task_text}{\textbf{\textcolor{task_text}{</task>}}}}

\newcommand{\taskresultStart}{\textboxStyle{task_text}{\textbf{\textcolor{task_text}{<result>}}}}
\newcommand{\taskresultEnd}{\textboxStyle{task_text}{\textbf{\textcolor{task_text}{</result>}}}}

\definecolor{bgcolor}{HTML}{F1F4F7}

\title{BrowseMaster: Towards Scalable Web Browsing via Tool-Augmented Programmatic Agent Pair}

\author{%
  \textbf{Xianghe Pang\textsuperscript{*} \quad Shuo Tang\textsuperscript{*} \quad Rui Ye \quad Yuwen Du \quad Yaxin Du \quad Siheng Chen\textsuperscript{†}}\\
  School of Artificial Intelligence, Shanghai Jiao Tong University\\
  \href{https://github.com/sjtu-sai-agents/Browse-Master}{https://github.com/sjtu-sai-agents/BrowseMaster}
}

\begin{document}

\maketitle
\renewcommand{\thefootnote}{}
\footnotetext{\textsuperscript{*} Equal contributions. \textsuperscript{†} Corresponding author: sihengc@sjtu.edu.cn.}
\renewcommand{\thefootnote}{\arabic{footnote}}

\begin{figure}[h]
    \centering
    \vspace{-3mm}
    \includegraphics[width=\linewidth]{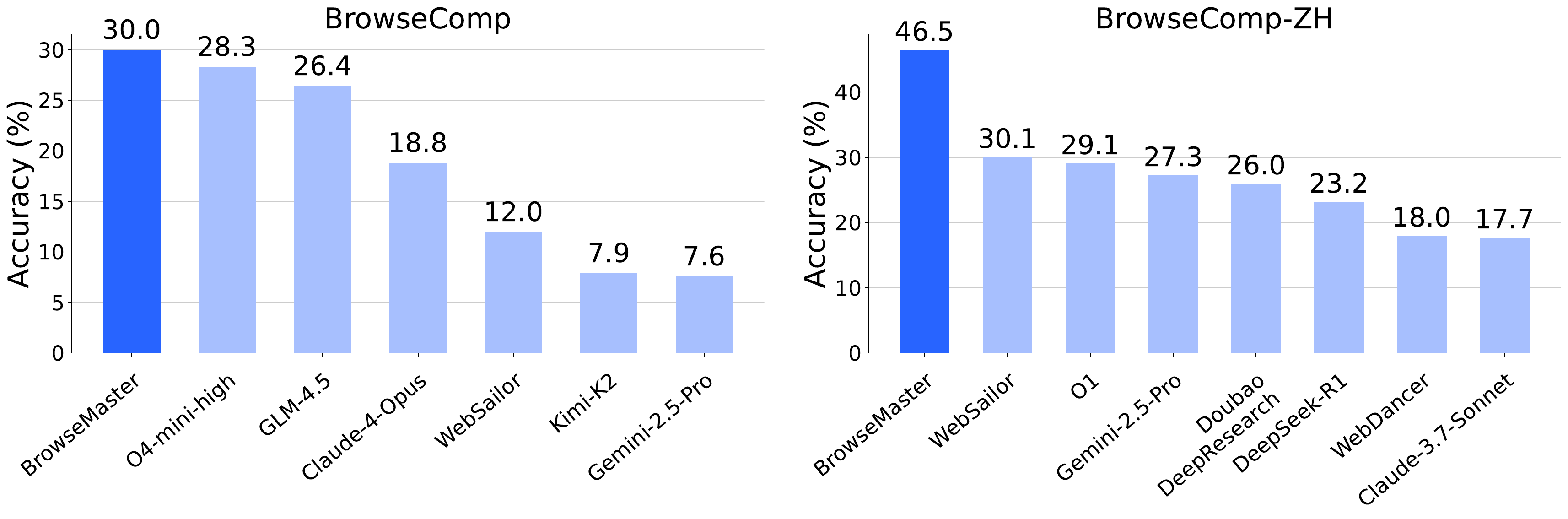}
    \caption{Comparisons on BrowseComp. Our \emph{BrowseMaster} achieves the score of \textbf{30\%}, surpassing deep research products from other baselines.}
    \label{fig:main_hle}
\end{figure}

\begin{abstract}

Effective information seeking in the vast and ever-growing digital landscape requires balancing expansive search with strategic reasoning. Current large language model (LLM)-based agents struggle to achieve this balance due to limitations in search breadth and reasoning depth, where slow, serial querying restricts coverage of relevant sources and noisy raw inputs disrupt the continuity of multi-step reasoning. To address these challenges, we propose BrowseMaster, a scalable framework built around a programmatically augmented planner-executor agent pair. The planner formulates and adapts search strategies based on task constraints, while the executor conducts efficient, targeted retrieval to supply the planner with concise, relevant evidence.  This division of labor preserves coherent, long-horizon reasoning while sustaining broad and systematic exploration, overcoming the trade-off that limits existing agents. Extensive experiments on challenging English and Chinese benchmarks show that BrowseMaster consistently outperforms open-source and proprietary baselines, achieving scores of 30.0 on BrowseComp-en and 46.5 on BrowseComp-zh, which demonstrates its strong capability in complex, reasoning-heavy information-seeking tasks at scale.

\end{abstract}

% \newpage
\section{Introduction}

Information seeking has been the engine of human progress, fueling discovery, shaping collective knowledge, and steering societal evolution~\citep{marchionini1995information,given2023looking}.
The advent of search engines (e.g., Google Search~\citep{brin1998anatomy}) constituted a paradigm shift, replacing slow, geographically constrained exploration with instantaneous, large-scale access to the world’s digitized knowledge.
Now, the rise of large language model (LLM)-based agents (e.g., Deep Research  from OpenAI~\citep{openai_deep_research}) signals the next revolution: systems capable of autonomously and tirelessly retrieving, synthesizing, and reasoning over web information—transcending the cognitive and operational limits of humans' search and charting a path toward automated information seeking.

% Information seeking, a hallmark of intelligent behavior, has driven human progress for centuries.
% Since its inception in 1998, Google have revolutionized information retrieval by indexing the web and delivering results in milliseconds. 
% However, traditional search systems burden users with navigating fragmented results, a process increasingly overwhelmed by the exponential growth of online content.
% Human cognitive limitations, such as bounded knowledge, distractibility, and sequential attention, make large-scale search challenging.
% Large Language Model (LLM) based agents present a promising paradigm shift, mitigating these limitations by autonomously issuing queries, browsing web pages, and reasoning across diverse sources. Emerging systems like DeepResearch fuel excitement in this domain, showcasing the potential for expansive web exploration and multi-step reasoning.

Effective information seeking requires reasoning to formulate precise search strategies and breadth to ensure comprehensive coverage of relevant information.
For example, \textit{identifying the title of a 2018–2023 EMNLP paper whose first author studied at Dartmouth College and whose fourth author studied at the University of Pennsylvania} demands reasoning over these constraints to devise an efficient search plan, while sustaining broad exploration to avoid missing the correct result.
Without sufficient reasoning, the process devolves into brute-force examination of thousands of papers; without sufficient breadth, it risks prematurely excluding the correct target.
By uniting strategic reasoning with expansive search, agents can tackle such tasks both effectively and at scale.

% Effective information seeking requires agents to synergistically balance search breadth and reasoning depth to address complex tasks. For example, consider the task of \textit{identifying the title of a 2018–2023 EMNLP paper where the first author studied at Dartmouth College and the fourth author studied at the University of Pennsylvania}. Broad search capabilities are essential, as they enable the exploration of a vast corpus of papers to locate relevant candidates. However, relying solely on broad searches is insufficient. A brute-force approach—sifting through over 3,000 EMNLP papers and manually checking author affiliations—would be excessively time-consuming and impractical. Instead, effective information seeking arises from a harmonious interplay between expansive search and deep reasoning. Deep reasoning is crucial for defining a concise, targeted search scope, while broad search ensures thorough and persistent exploration within that scope. This iterative cycle of scoping and searching underscores the need for information-seeking frameworks that seamlessly integrate these complementary strategies.

However, current LLM-based agents, remain constrained in their ability to combine expansive search with strategic reasoning.
First, their search breadth is limited: most invoke web browsing tools via natural language and process queries serially, resulting in a one-page-at-a-time workflow that drastically reduces the number of sources examined and undermines comprehensive coverage~\citep{webdancer}.
Second, their reasoning depth is shallow: each tool invocation injects raw web content into the agent’s context, interrupting the flow of reasoning and fragmenting multi-step inference~\citep{li2025websailor}.
These limitations, acting in tandem, lead to near-zero accuracy on challenging information-seeking tasks~\citep{search-o1,search-r1,webthinker}, highlighting the urgent need for architectures that can maintain broad exploration while preserving coherent reasoning.

% Current LLM-based agents, despite their potential, struggle to balance search breadth and reasoning depth due to design limitations.
% First, insufficient search breadth arises as agents invoke tools through natural language, processing queries serially. This one-page-at-a-time browsing severely restricts the number of sources an agent can explore, hindering comprehensive coverage. Second, shallow reasoning depth emerges because execution disrupts reasoning continuity. Each tool call introduces raw web content into the agent’s context, which fragments multi-step reasoning and impedes sustained focus. Overall, these constraints cause agents to exhibit near-zero accuracy on challenging information-seeking tasks, underscoring the urgent need for designs that enable efficient search and robust reasoning.

To address the limitations in achieving both search breadth and reasoning depth, we present BrowseMaster, a framework for scalable, reasoning-intensive web browsing built around a tightly coordinated planner–executor agent pair.
In our design, the planner focuses on high-level reasoning, formulating strategies and delegating well-defined sub-tasks to the executor;
while the executor concentrates on executing these tasks through multi-step interactions with the environment.
This separation keeps the planner’s context clean, shielding its reasoning process from noisy environmental outputs, and allows the executor to remain fully engaged with sub-task execution and high-volume interactions.

The two components in BrowseMaster play distinct yet complementary roles:
(1) Planner: long-horizon strategist.
The planner interprets the task, extracts key constraints, and formulates a search strategy that incrementally refines the problem space.
Operating solely over structured outputs returned by the executor, it avoids the fragmentation of multi-step reasoning caused by direct exposure to raw, unprocessed web content.
It further employs confidence-guided replanning, which resets its context and revises the strategy when confidence is low, thus preventing premature convergence and enabling adaptive reasoning over extended horizons.
(2) Executor: scalable search engine.
The executor enables expansive, efficient search at scale by interacting with tools programmatically, representing operations such as search, parse, and check as composable code primitives.
This design allows selective extraction of relevant information (e.g., printing only pertinent pages), drastically reducing context size and processing overhead.
By encoding complex search workflows in compact code, the executor can sustain a high volume of environment interactions without overloading the reasoning proces, overcoming the inefficiencies of prior agents that rely on slow, natural-language tool calls.
Together, the planner maintains coherent reasoning while the executor ensures broad, systematic exploration, enabling BrowseMaster to achieve scalable and effective information seeking.

Experimentally, we evaluate BrowseMaster on challenging web browsing benchmarks covering both English and Chinese tasks, against a range of open-source and proprietary agents.
Results demonstrate that BrowseMaster leverages creative, code-based search strategies to efficiently navigate thousands of pages and reason effectively over diverse search cues, consistently delivering strong performance on long-horizon, information-rich tasks.
On BrowseComp-en~\citep{wei2025browsecomp}, it achieves a score of 30.0, becoming the first open-source agent to reach this milestone.
On BrowseComp-zh~\citep{zhou2025browsecomp}, it surpasses OpenAI’s DeepResearch~\citep{openai_deep_research} by 4\% and outperforms other advanced proprietary models such as o1~\citep{o1-preview} and Doubao~\citep{doubao2025}.

% Experimentally, we comprehensively demonstrate the effectiveness of BrowseMaster across challenging web browsing benchmarks, spanning both English and Chinese tasks, and comparing against both open-source and proprietary agents. Results show that BrowseMaster employs creative code-based search strategies, efficiently navigates through thousands of pages, and reasons effectively using diverse search cues. 
% As a result, BrowseMaster consistently delivers strong search performance on long-horizon, information-rich tasks.
% Notably, BrowseMaster achieves a score of 30.0 on BrowseComp-en, becoming the first open-source agent to reach this milestone. On BrowseComp-zh, it surpasses OpenAI’s DeepResearch by 4\% and outperforms advanced proprietary models such as o1 and Doubao.

\section{Planner-Executor Agent Pair} \label{sec:agent}

This section presents the design of our \emph{Planner-Executor Agent Pair}, beginning with an overview, followed by the design of the planner and executor components.

\begin{figure}[h]
    \centering
    \includegraphics[width=\linewidth]{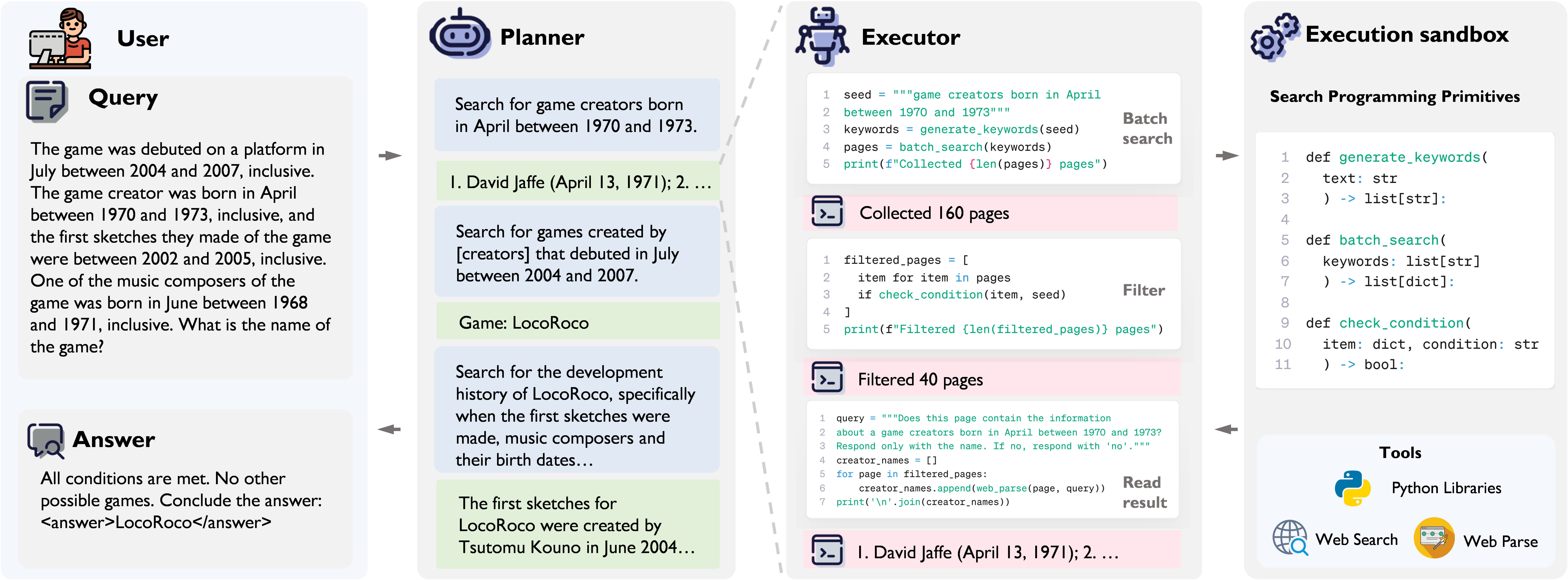}
    \caption{The architecture of \emph{BrowseMaster}.}
    \label{fig:structure}
\end{figure}

\subsection{Workflow Overview}
Our workflow primarily focuses on providing a more efficient context management mechanism to further enhance the search breadth and the reasoning depth during agent browsing. This improvement targets two key performance dimensions: 
1) Complex reasoning and plannin, the agent must adapt search strategies dynamically by leveraging diverse clues encountered during browsing;
2) Execution capability – the agent must sustain a high volume of tool calls to gather necessary information, while detecting and recovering from tool failures or network issues.
To this end, we extend the standard ReAct architecture by introducing two specialized agents (Sections~\ref{subsec:master} and~\ref{subsec:worker}): a planner responsible for strategic reasoning and planning, and an executor responsible for tool-augmented task execution.
In each execution cycle, the planner processes the user query, performs reasoning, and decomposes the task into subtasks.
Information retrieval subtasks are delegated to the executor, which interacts with tools programmatically. 
Through a sequence of tool invocations, the executor produces distilled intermediate results and returns them to the planner for coordination and integration.

% To achieve this, we modify the standard ReAct architecture by introducing two specialized agents (detailed in Sections~\ref{subsec:master} and~\ref{subsec:worker}). The planner agent focuses on strategic reasoning and planning, while the executor agent executes tasks using tools. In each execution cycle, the user query is first processed by the planner agent, which performs reasoning and decomposes the task into subtasks. Information retrieval subtasks are delegated to the executor agent, which executes the assigned operations. Through a series of complex tool invocations, the executor agent derives the final result and returns it to the planner for further coordination and integration. 

This design offers two main advantages.
(1) Preserving reasoning depth. By isolating tool execution from the planner, we prevent noisy execution details from disrupting multi-step inference.
(2) Expanding search breadth. By delegating well-defined subtasks, the executor can perform searches that are both more targeted and more extensive.
The overall architecture is illustrated in Figure~\ref{fig:structure}.

% This design offers two benefits: 1) For reasoning, it prevents tool execution details from disrupting the planner’s inference, preserving reasoning depth. 2) For tool execution, it eliminates the need to carry the full reasoning context, enabling more frequent searches and expanding search breadth; see Figure~\ref{fig:structure} for the architecture.

\subsection{Planner: Confidence-Guided Replanning for Persistent Exploration}
\label{subsec:master}

The planner performs long-horizon reasoning over the input search task by decomposing it into manageable sub-tasks and delegating their execution to the executor.
During reasoning, the planner invokes the executor by enclosing the assigned sub-task within a \taskStart \taskEnd block.
Upon completion, the executor’s outputs are inserted into the \taskresultStart \taskresultEnd block, after which the planner continues reasoning with its updated context.
To enhance inference-time scalability, the planner produces a confidence score when arriving at a final answer; if the score is below a predefined threshold, it triggers replanning to refine the solution.

% The planner is responsible for performing long-term reasoning over the input search task by decomposing complex tasks into more manageable sub-tasks and delegating them to a worker for execution. Inspired by the React paradigm, we introduce two special tokens: \taskStart \taskEnd and \taskresultStart \taskresultEnd. During the reasoning process, the planner inserts the generated sub-tasks into the \taskStart \taskEnd block and invokes the worker to execute them. Once the worker completes these tasks and returns the results, they are inserted into the \taskresultStart \taskresultEnd block. The model then continues reasoning based on this updated context. To further support inference-time scalability, the planner is also required to generate a confidence score upon reaching a final answer. If the score falls below a predefined threshold, the planner will replan accordingly. 

Here, the planner is driven by a reasoning model, leveraging the model's inherent logical reasoning capabilities to analyze and decompose complex tasks, rather than relying on a fixed workflow.

\subsection{Executor: Tool-Augmented Browse Worker Mechanism}
\label{subsec:worker}

The executor is responsible for maximizing both the quantity and quality of tool calls to collect as much accurate and relevant information as possible for the planner.
Since task decomposition is handled by the planner, the executor’s role is not to break down tasks, but to explore unsearched aspects of the information space.
Its behavior is therefore primarily operational, involving systematic web browsing, information gathering, and refinement.
To ensure efficient and comprehensive information collection, the executor incorporates the following key components:

% Here, the executor focuses on improving the quantity and quality of tool calls in order to collect as much information as possible for the planner. Since task decomposition is guided by the planner, what the executor needs to do is not task breakdown, but rather to explore aspects that have not yet been searched. In this context, the agent’s primary behavior is more mechanical, involving browsing web pages, gathering information, and refining it. To efficiently and comprehensively collect information, we have designed the executor with the following key elements:

\noindent\textbf{Using code execution as interaction.}
We enable the model to invoke tools by generating executable code within  \codeStart \codeEnd tags.
The extracted code segment, identified via matching rules, is executed in a sandboxed environment with the relevant tool functions pre-imported.
Execution outputs are then wrapped in \resultStart \resultEnd tags and appended to the model’s context, allowing inference to continue seamlessly.
Details of the available tools and execution environment are provided in Sections~\ref{subsec:tools} and~\ref{subsec:exe_env}.

% Specifically, we allow the model to write code to call tools. The tool-call code written by the model is placed inside \codeStart \codeEnd tags in the prompt. Once the code segment is extracted based on matching rules, it is executed in a code execution sandbox, where various tool functions are imported. We then place the execution results in the \resultStart \resultEnd tags and let the model continue inference; see details in Section~\ref{subsec:tools} and Section~\ref{subsec:exe_env} for the tools and execution environment.

\noindent\textbf{Standardized search programming primitives.}
Just as Python ships with a rich standard library to encapsulate common operations, web search agents can benefit from built-in, task-specific primitives. In large-scale information seeking, certain patterns recur frequently—such as expanding a query with multiple keyword variants or verifying whether a retrieved page contains target information.

Without such primitives, these steps must be reimplemented from scratch, causing redundancy and a higher risk of errors. Abstracting them into \emph{modular, reusable functions} that encapsulate common search behaviors gives the agent a stable, high-level API for tool interaction.

This design offers two main benefits:
i) reducing redundancy, as the same primitive can serve diverse tasks without rewriting low-level logic; and
ii) improving flexibility and scalability, as primitives can be composed or customized to dynamically refine search strategies. Overall, encapsulating search logic in such modular units enables efficient, adaptable, and extensible web exploration.

\section{Tool-Augmented Programmatic Sandbox}

To equip the executor with reliable and expressive tool-use capabilities, we introduce the \emph{tool-augmented programmatic sandbox}, a unified framework for precise and controllable interaction with the external environment.
The sandbox exposes standardized programming primitives tailored for web-based tasks and supports code execution within a lightweight, isolated runtime.
It serves as the execution backbone of our agent, translating the planner’s strategic intent into actionable and verifiable operations.

\subsection{Standardized Search Programming Primitives}
\label{subsec:primitive}
In web search tasks, procedural control structures (e.g., loops and conditional branches) can substantially improve execution efficiency.
For example, a single code execution may generate numerous search queries, perform concurrent retrieval via multithreading, and filter the results according to unified rules.
However, directly prompting the model to write complete control code often leads to instability: webpages differ widely in format and structure, making it challenging to implement universal filtering strategies.
As a result, generated code frequently fails in handling corner cases, causing wasted time on debugging and error correction.

To address this, we design a set of standardized programming primitives specifically for agent-based web search:
\colorbox{result}{\texttt{generate\_keywords}}, \colorbox{result}{\texttt{batch\_search}}, and \colorbox{result}{\texttt{check\_condition}}.
These encapsulate the key capabilities of generating search queries, performing parallel retrieval, and applying programmable filtering logic.

\noindent\colorbox{result}{\textbf{\texttt{generate\_keywords(seed\_keyword)}}}
generates a set of search terms starting from a seed keyword, producing advanced search expressions such as conditional filters or domain-specific queries (e.g., restricting to Wikipedia).
The goal is to broaden coverage and capture semantically related content that may not be retrieved with a single query.

% \noindent \colorbox{result}{\textbf{\texttt{generate\_keywords(seed\_key\_word)}}} The \texttt{generate\_keywords} primitive is used to produce a set of search terms for web queries. Starting from a seed keyword, it generates a series of advanced search expressions suitable for search engines—such as conditional filters or domain-specific searches (e.g., targeting Wikipedia)—with the goal of retrieving more comprehensive information related to the original keyword.

\noindent \colorbox{result}{\textbf{\texttt{batch\_search(key\_words)}}}
executes multiple web searches in parallel, substantially improving efficiency over traditional step-by-step querying.
Rather than issuing individual search requests sequentially, the agent can submit an entire batch of queries simultaneously and receive all results in a single step.
The input is a list of search keywords, either generated directly by the agent or derived from the output of \texttt{generate\_keywords}.
This parallel execution enables the agent to retrieve information from a large number of webpages quickly, while maintaining both coverage and speed.

\noindent \colorbox{result}{\textbf{\texttt{check\_condition(web\_page, condition)}}}
In large-scale web search, agents must process and analyze substantial volumes of information, making efficient filtering and conditional evaluation essential.
The \texttt{check\_condition} primitive offers a programmable interface for code-driven, large-scale content evaluation, replacing slow, sequential manual inspection by the model.
It accepts two inputs:
(1) a batch of document contents (e.g., webpage text), and
(2) a declarative condition expressed as a model-generated predicate or logical statement.
It returns a Boolean value for each input—\texttt{True} if the condition is met, and \texttt{False} if it is not satisfied or cannot be determined from the content.
By leveraging \texttt{check\_condition}, agents can construct efficient, logic-based filtering pipelines and make control-flow decisions grounded in semantic conditions.
This abstraction supports scalable post-processing of web data and fine-grained control over downstream decision-making, all within a code-executed framework.

% In large-scale web search tasks, agents often need to process and analyze vast amounts of information. Efficient filtering and conditional evaluation over this data is a core requirement. The \texttt{check\_condition} primitive provides a programmable interface for conditional filtering, enabling agents to evaluate content at scale using code-driven logic, rather than relying on sequential, manual inspection by the model. This primitive accepts two inputs: 1) A batch of document contents (e.g., web page text). 2) A declarative condition expressed as a model-generated predicate or logical statement. It returns a boolean output for each input—True if the condition is satisfied, and False if the condition is not met or cannot be determined from the given content. By leveraging \texttt{check\_condition}, agents can implement efficient, logic-based filtering pipelines and control flow decisions based on semantic conditions. This abstraction allows for scalable post-processing of web data and fine-grained control over downstream decision-making, all within a code-executed framework.

By using these structured functions, the model can write more reliable and maintainable code, significantly improving execution stability and reducing implementation complexity.

\subsection{Tools}
\label{subsec:tools}

To mimic human-like online information-seeking, we design two essential tools: web search and web parse. The web search tool empowers the agent to identify relevant web pages based on the question.
It delivers concise summaries for each retrieved page, allowing the agent to strategically determine which links warrant deeper exploration.
The web parse tool is employed when the agent requires in-depth analysis of a selected webpage to extract information directly related to the user query.

\textbf{Web search.}
The web search tool utilizes Google search engine to pinpoint the most relevant webpages based on a user's query. It delivers three key categories of valuable information:
(i) Entity-related facts: 
For queries involving recognizable entities (such as a company or software application), the tool identifies them and pulls structured facts from its knowledge graph. This includes the entity's name, a brief description, and essential attributes. By extracting these details, the agent can quickly grasp the query's central concept, offering vital context for further analysis.
(ii) Relevant webpage previews:
For each matching page, the tool supplies a preview that includes the title, URL, and an informative snippet. This allows the agent to rapidly evaluate the page's relevance and decide which ones merit closer inspection.
(iii) Related search queries: The tool also suggests common follow-up searches, giving the agent options to refine or expand the investigation and foster a more comprehensive grasp of the topic.

\textbf{Web parse.}
The web parse tool supports two specialized parsing approaches, one for standard webpages and another for scientific papers:
(i) General webpage parsing: 
This strategy starts by extracting the main content from the target webpage. 
To ensure robust operation, a fallback mechanism is incorporated to manage instances where direct content extraction fails. 
Once the content is obtained, the tool highlights sections most pertinent to the query. It also automatically identifies and lists links to related subpages, complete with short descriptions. This capability lets the agent delve deeper into connected content, mimicking human web navigation—scanning links, following trails, and building a fuller picture of the topic.
(ii) Scientific paper parsing: For scientific papers, the tool uses a two-step strategy to ensure reliable content retrieval. 
It first attempts to fetch an HTML version of the publication from ar5iv.
In the event of an unsuccessful or incomplete HTML fetch, the system switches to downloading the official PDF. With the full document in hand, the tool then extracts details directly tied to the query.

Together, the web search and web parse tools empower the agent not just to locate key information, but to navigate the web in a natural, human-inspired manner—through iterative searching, previewing, linking, and in-depth exploration as required.

\subsection{Execution Environment}
\label{subsec:exe_env}

We enable agents to invoke tools through code generation.
However, conventional stateless code execution sandboxes are poorly suited for multi-step tool use, as agents often define functions or variables in earlier code blocks and reference them later.
In a stateless sandbox, each execution occurs in an isolated memory space, preventing access to previously defined entities and severely restricting coding flexibility.

To overcome this limitation, we design a stateful code execution sandbox.
Each agent is allocated an isolated execution environment with persistent memory, allowing the execution state to be preserved and restored between runs.
This design offers a Jupyter Notebook–like experience, enabling agents to flexibly define and reuse functions, classes, and objects across multiple steps.
Meanwhile, different queries are executed in fully isolated contexts, ensuring clean separation and preventing cross-task interference.

% We enable agents to call tools by writing code. However, traditional stateless code execution sandboxes are not well-suited for multi-step tool use by agents. This is because agents often define functions or variables in earlier code blocks and reference them in later ones. In a typical stateless sandbox, each code execution runs in an isolated memory space, making it impossible to access previously defined functions or objects. This severely limits the flexibility of agents when writing code.

% To address this limitation, we developed a stateful code execution sandbox. Each agent is assigned an isolated execution environment with persistent memory. By capturing and restoring the execution state between runs, we provide a Jupyter Notebook–like experience where agents can flexibly define and reuse functions, classes, and objects across multiple steps. At the same time, different queries are executed in fully isolated contexts to ensure clean separation and avoid interference.

\section{Experiments}

\subsection{Experimental Setups}
\textbf{Agent.}
We employ DeepSeek-R1-0528~\citep{deepseek-r1-0528} to drive the planner and DeepSeek-R1 for the executor.
The maximum completion of tokens is set to 64k with a temperature of 0.6.

\textbf{Benchmarks.}
We evaluate our method on five challenging benchmarks: BrowseComp~\citep{wei2025browsecomp}, a highly demanding benchmark designed to assess the ability to locate complex, entangled information; BrowseComp-zh~\citep{zhou2025browsecomp}, a Chinese counterpart to BrowseComp with similar objectives; xBench-DeepResearch~\citep{chen2025xbench}, a dynamic benchmark focused on evaluating tool usage in search and information retrieval tasks; GAIA~\citep{mialon2023gaia}, which tests reasoning, web browsing, and general tool-use capabilities; and WebWalkerQA~\citep{wu2025webwalker}, which assesses agents' ability to navigate and process complex, multi-layered web information.

Due to resource constraints of the search API, we randomly sample 200 examples for BrowseComp and BrowseComp-zh. For GAIA, we use the text-only queries from its validation set (103 samples). Evaluation employs xVerify-9B~\citep{chen2025xverify} for BrowseComp, BrowseComp-zh, and xBench-DeepResearch, GPT-4o~\citep{gpt-4o} for WebWalkerQA and GAIA following~\cite{wu2025webwalker}.

\textbf{Baselines.} We compare our performance against systems from three categories: proprietary deep research agents (OpenAI~\citep{openai_deep_research}, Gemini 2.5~\citep{geminideepresearch}, Grok3~\citep{xai2025grok3}, Doubao~\citep{doubao2025}, and Metaso~\citep{metaso2025}), advanced models (QwQ~\citep{qwen2025qwq32b}, DeepSeek-R1~\citep{deepseek-r1-0528}, GPT-4o~\citep{gpt-4o}, Gemini 2.5 Pro~\citep{gemini-2.5-pro}, and o1~\citep{o1-preview}), and open-source agents (WebThinker~\citep{webthinker}, WebDancer~\citep{webdancer}, WebSailor~\citep{li2025websailor}, WebShaper~\citep{tao2025webshaper}, and Agentic Reasoning~\citep{agentic_reasoning}). Due to limited API access for proprietary agents and models, not all systems were evaluated across every benchmark. For open-source agents without full accessibility, we use their officially reported results from their respective papers~\citep{li2025websailor, tao2025webshaper}.

\begin{table}[t]
\centering
\caption{Performance comparison against proprietary agents, advanced models, and open-source agents on five benchmarks. BrowseMaster outperforms all open-source agents and advanced models, as well as most proprietary deep research agents.}
\resizebox{\textwidth}{!}{%
\begin{tabular}{lccccc}
\toprule
 & BrowseComp & BrowseComp-zh & xbench-DeepSearch & GAIA & WebWalkerQA \\
\midrule
\textbf{Proprietary Agents} \\
\quad OpenAI DeepResearch & \textbf{51.5} & 42.9 & - & \underline{67.4} & - \\
% \quad Gemini2.5 DeepResearch & & - & 50+ & - & - \\
\quad Grok3 DeepResearch & - & 12.9 & 50+ & - & - \\
\quad Doubao DeepResearch & - & 26.0 & 50+ & - & - \\
\quad Metaso DeepResearch & 12.0 & \underline{45.3} & \underline{64.0} & - & - \\
\midrule
\textbf{Models} \\
\quad QwQ & 0.5 & 10.0 & 10.7 & 22.3 & 4.3 \\
\quad DeepSeek-R1 & 2.0 & 23.2 & 32.7 & 16.5 & 10.0 \\
\quad GPT-4o & 0.6 & 6.2 & 18.0 & 17.5 & 5.5 \\
\quad Gemini 2.5 Pro & 7.6 & 27.3 & - & - & - \\
\quad OpenAI o1 & 9.9 & 29.1 & - & - & 9.9 \\
\midrule
\textbf{Open-source Agents} \\
\quad WebThinker & 1.5 & 7.3 & 24.0 & 48.5 & 39.4 \\
\quad WebDancer & 3.8 & 18.0 & 39.0 & 51.5 & 43.2 \\
\quad WebSailor & 12.0 & 30.1 & 55.0 & 55.4 & - \\
\quad WebShaper & - & - & - & 60.2 & 52.2 \\
\quad Agentic Reasoning & 5.5 & 29.0 & 40.0 & 42.2 & 36.9 \\
\quad BrowseMaster & \underline{30.0} & \textbf{46.5} & \textbf{66.0} & \textbf{68.0} & \textbf{62.1} \\
\bottomrule
\end{tabular}
}
\end{table}

\subsection{Main Results}

\textbf{BrowseMaster achieves superior performance over both open-source and proprietary agents.} As the first open-source agent to exceed a 30\% score, BrowseMaster represents a significant leap forward, showcasing the power of programmatic execution and agentic workflows. While leading deep research agents are typically proprietary,  BrowseMaster establishes an open-source paradigm for tackling challenging search tasks. Notably, it to outperform systems like Grok3 and Doubao DeepResearch and achieves a 4\% performance advantage over OpenAI's DeepResearch on BrowseComp-zh.

\textbf{BrowseMaster excels consistently across diverse benchmarks and languages.} BrowseMaster adaptively handles both complex search tasks like BrowseComp and web traversal challenges like WebWalkerQA in both Chinese and English, demonstrating its versatility. Performance gain is particularly impressive on deep research benchmarks, where persistent exploration and broad coverage are critical, underscoring BrowseMaster's exceptional design for search breadth and reasoning depth.

\textbf{Tool-augmented reasoning significantly boosts performance on information-seeking tasks.} Advanced standalone models like GPT-4o and DeepSeek-R1 achieve near-zero performance on BrowseComp, indicating that raw models struggle without web interaction. Equipped with web-browsing capabilities, BrowseMaster substantially enhances DeepSeek-R1's performance, surpassing proprietary models like Gemini 2.5 Pro and o1. By accessing, filtering, and reasoning over vast web data, BrowseMaster tackles real-world challenges unattainable by pure language models.

\subsection{Analysis} \label{sec:analysis}
\begin{figure}[h!]
    \centering
    \begin{minipage}[t]{0.61\textwidth}
        \centering
        \begin{subfigure}[t]{0.48\textwidth}
            \centering
            \includegraphics[width=\linewidth, height=5cm, keepaspectratio]{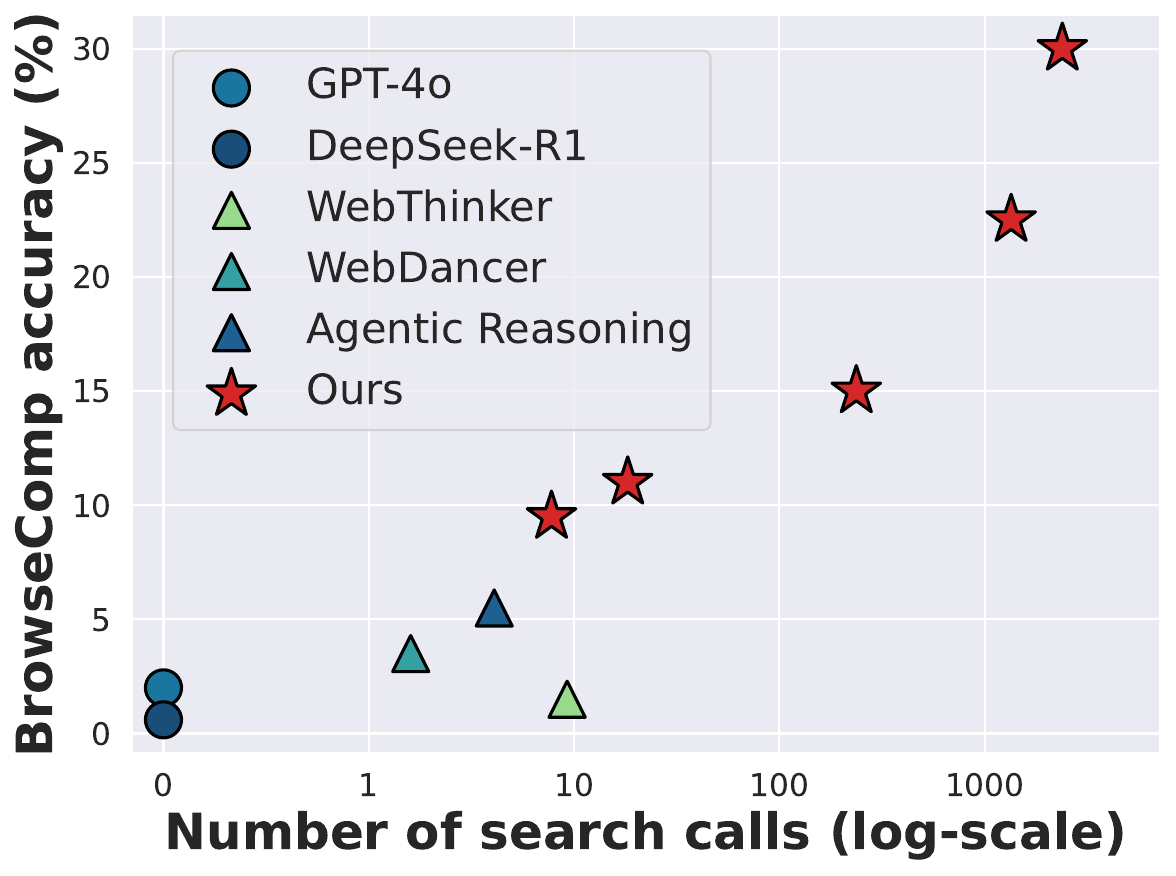}
            % \caption{Search call volume}
            \label{fig:scale_search}
        \end{subfigure}
        \hfill
        \begin{subfigure}[t]{0.48\textwidth}
            \centering
            \includegraphics[width=\linewidth, height=5cm, keepaspectratio]{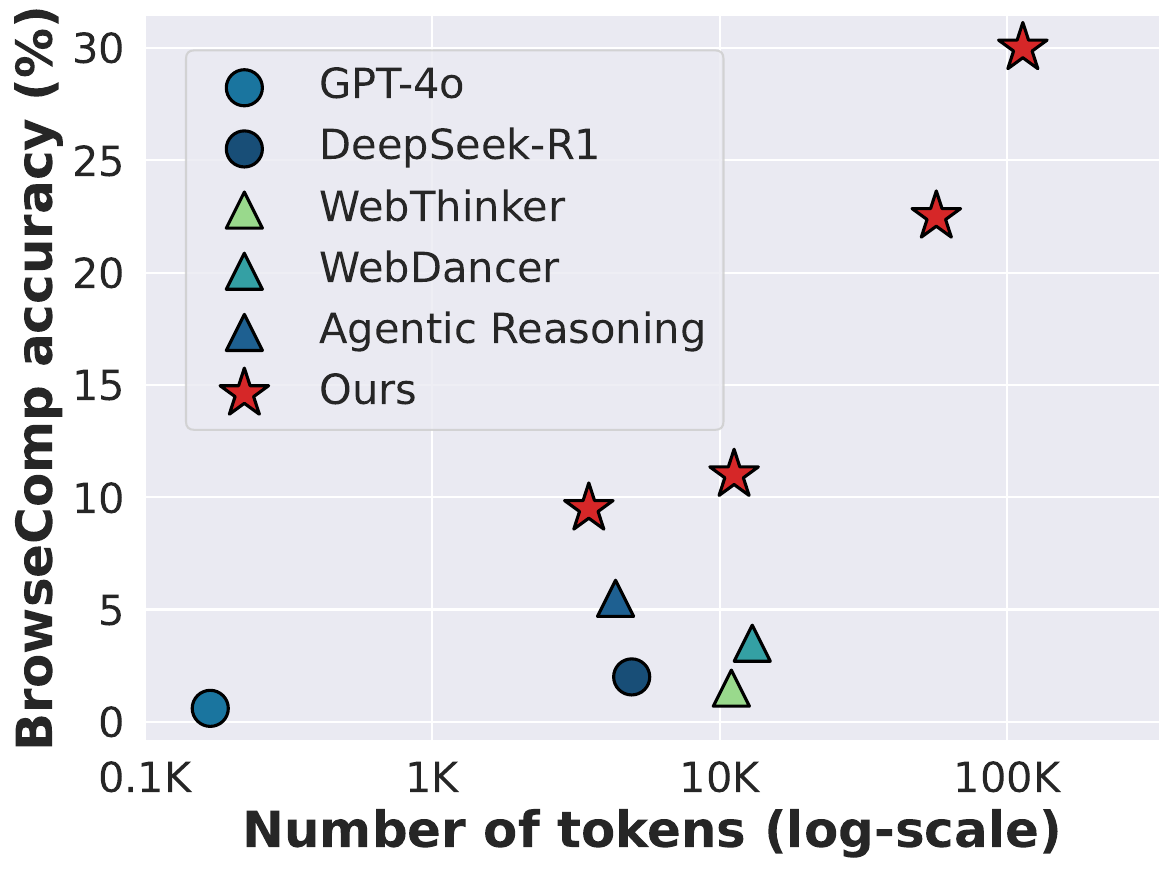}
            % \caption{Token usage}
            \label{fig:scale_computation}
        \end{subfigure}
        \vspace{0.5em} % 调整标题间距
        \caption{Performance comparison in terms of search call volume and total token usage. Scaling search calls and computation drives performance gains.}
        \label{fig:scale_search_and_think}
    \end{minipage}
    \hfill
    \begin{minipage}[t]{0.37\textwidth}
        \centering
        \includegraphics[width=\linewidth, height=5cm, keepaspectratio]{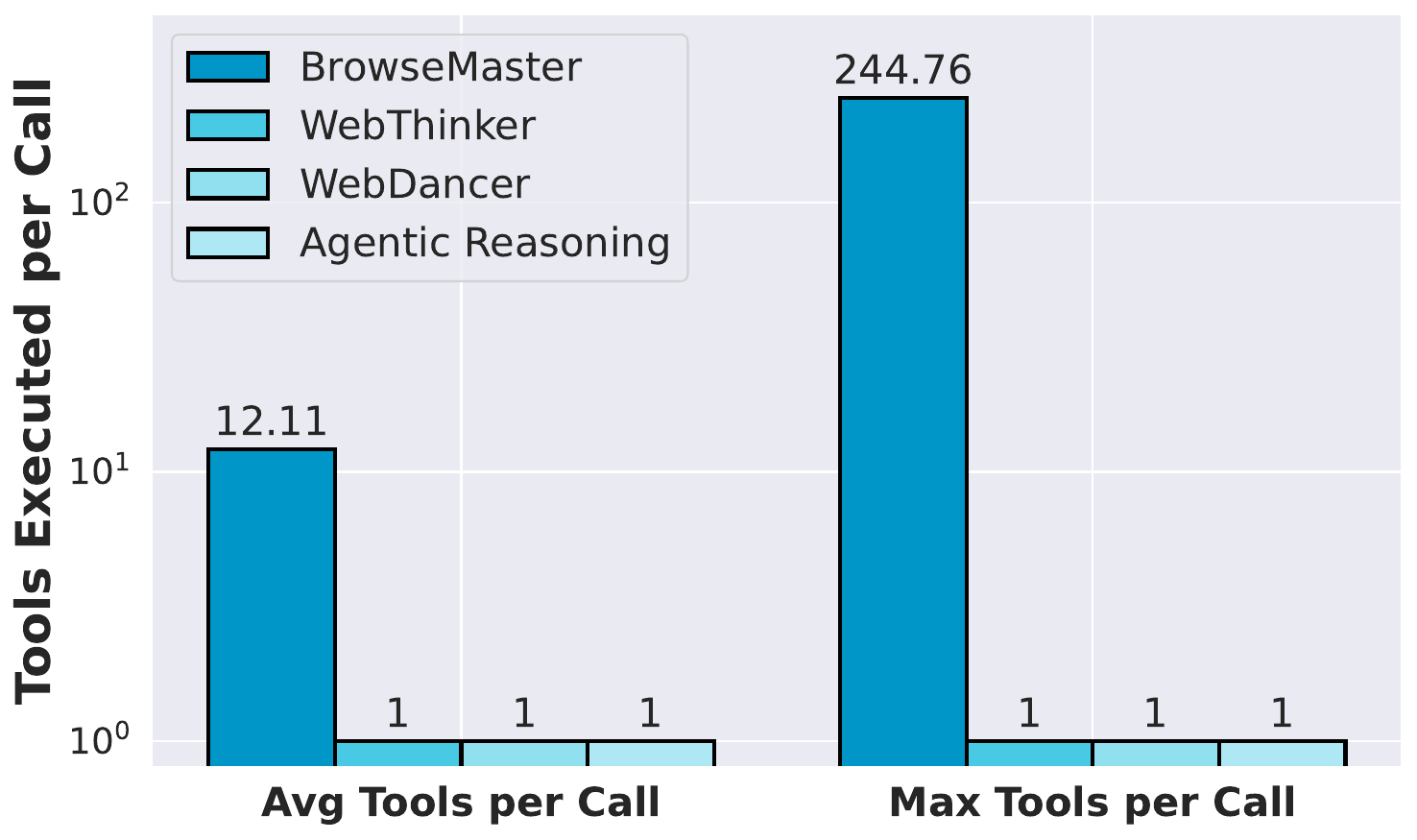}
        \vspace{0.5em} % 匹配左侧间距
        \caption{Comparison of tool calls per invocation. Code-driven execution enables highly efficient tool calls.}
        \label{fig:tool_call_compare}
    \end{minipage}
\end{figure}

% \begin{figure*}[h!]
%     \centering
%     \begin{subfigure}[b]{0.45\textwidth}
%         \centering
%         \includegraphics[width=\linewidth]{figs/search_score_plot_woDR.pdf}
%         \caption{Performance comparison in terms of search call volume. Scaling search calls drives performance gains.}
%         \label{fig:scale_search}
%     \end{subfigure}
%     % \hspace{0.15cm}
%     \begin{subfigure}[b]{0.45\textwidth}
%         \centering
%         \includegraphics[width=\linewidth]{figs/token_score_plot_woDR.pdf}
%         \caption{Performance comparison in terms of total token usage. Increasing computation drives performance gains.}
%         \label{fig:scale_computation}
%     \end{subfigure}
%     % \caption{}
%     % \vspace{-.5em}
%     % \caption{Analysis of the scale of agents and scenarios.}
%     \label{fig:scale_search_and_think}
%     % \vspace{-.6em}
% \end{figure*}

% \begin{figure}
%     \centering
%     \includegraphics[width=0.5\linewidth]{figs/tool_call_compare.pdf}
%     \caption{Comparison of tool calls per invocation between BrowseMaster and WebThinker on BrowseComp. BrowseMaster’s  code-driven approach enables significantly more efficient tool calls.}
%     \label{fig:tool_call_compare}
% \end{figure}

\textbf{Scaling search calls empowers BrowseMaster to achieve performance breakthrough.} Figure~\ref{fig:scale_search_and_think} illustrates the performance of BrowseMaster and baselines on BrowseComp as a function of search call. We evaluate BrowseMaster across configurations combining the executor with and without primitives and planner. The results show that i) at equivalent search call levels, BrowseMaster surpasses existing open-source agents; ii) scaling search call volume is critical for enhancing agent performance, as relying on fewer than 10 searches is often impractical for challenging search tasks; and iii) BrowseMaster’s search capabilities significantly enhance the performance of the pure model.

\textbf{Scaling computation empowers BrowseMaster to achieve performance breakthrough.} Figure~\ref{fig:scale_search_and_think} illustrates the performance of BrowseMaster and baselines on BrowseComp as a function of total token usage. The results demonstrate that BrowseMaster significantly enhances agent performance by scaling computation. This scaling arises from the synergistic collaboration between the planner and executor. The planner decomposes complex problems into manageable subtasks, allowing the executor to tackle lower-difficulty tasks incrementally, progressively solving the overall problem. Increased computational resources enable BrowseMaster to reason deeply, connect clues, optimize search directions, and validate results.

\begin{figure}[h!]
    \centering
    \begin{minipage}[t]{0.42\textwidth}
        \centering
        \includegraphics[width=\linewidth, keepaspectratio]{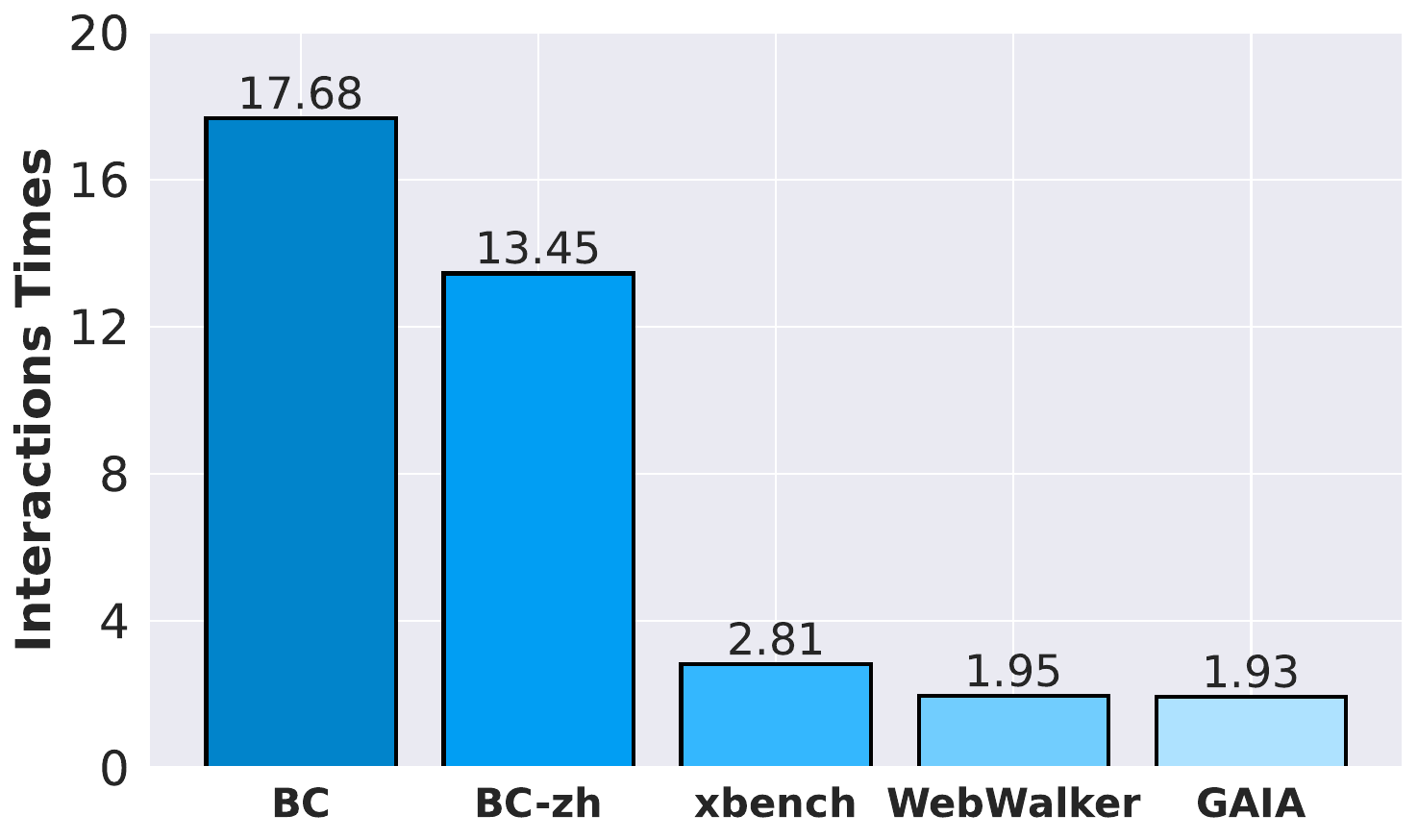}
        \caption{Interaction times between planner and executor across benchmarks. Complex tasks require increased task decomposition and confidence-guided retries.}
        \label{fig:interaction_times}
    \end{minipage}
    \hfill
    \begin{minipage}[t]{0.54\textwidth}
        \centering
        \begin{subfigure}[t]{0.48\textwidth}
            \centering
            \includegraphics[width=\linewidth, keepaspectratio]{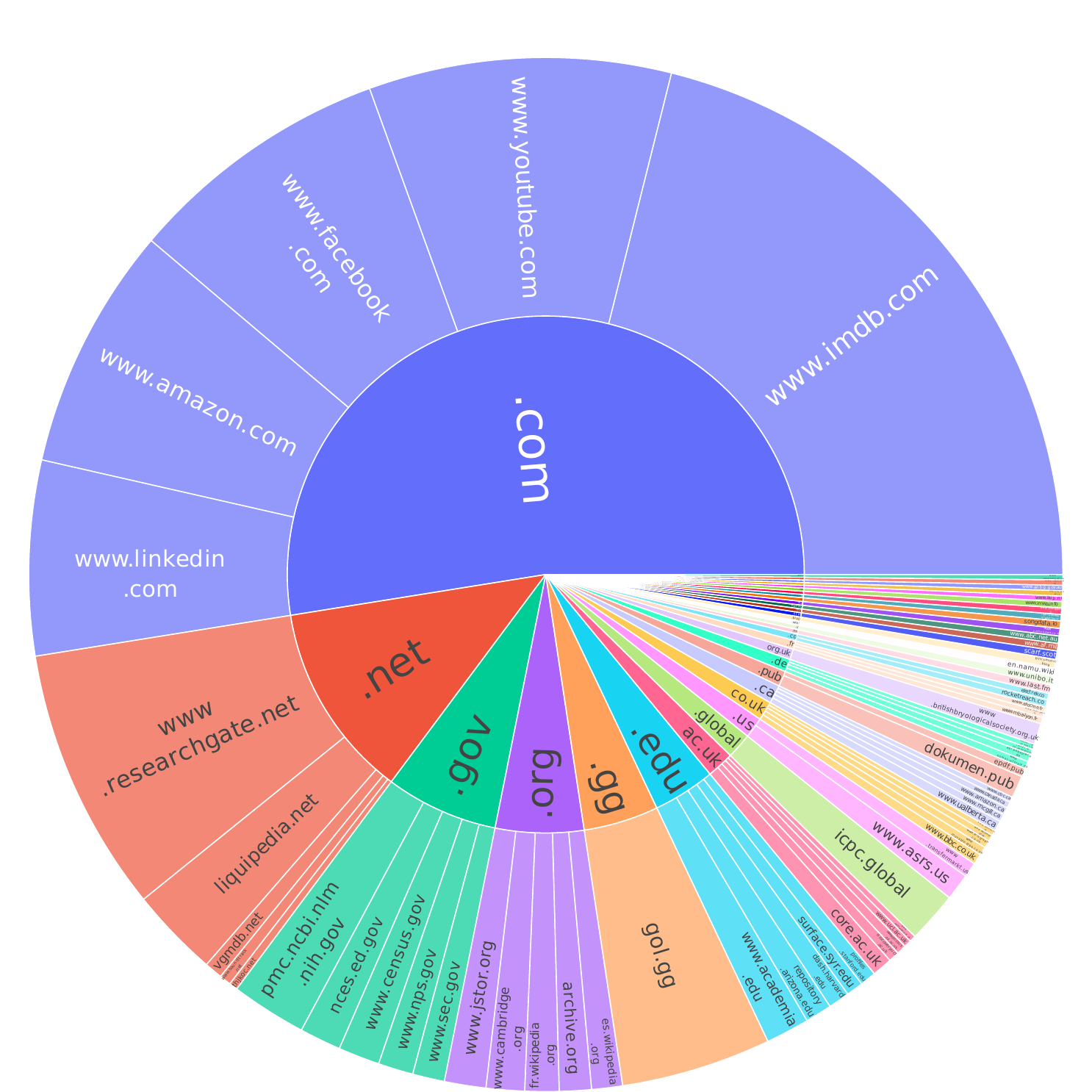}
            \caption{BrowseMaster.}
        \end{subfigure}
        \hfill
        \begin{subfigure}[t]{0.48\textwidth}
            \centering
            \includegraphics[width=\linewidth, keepaspectratio]{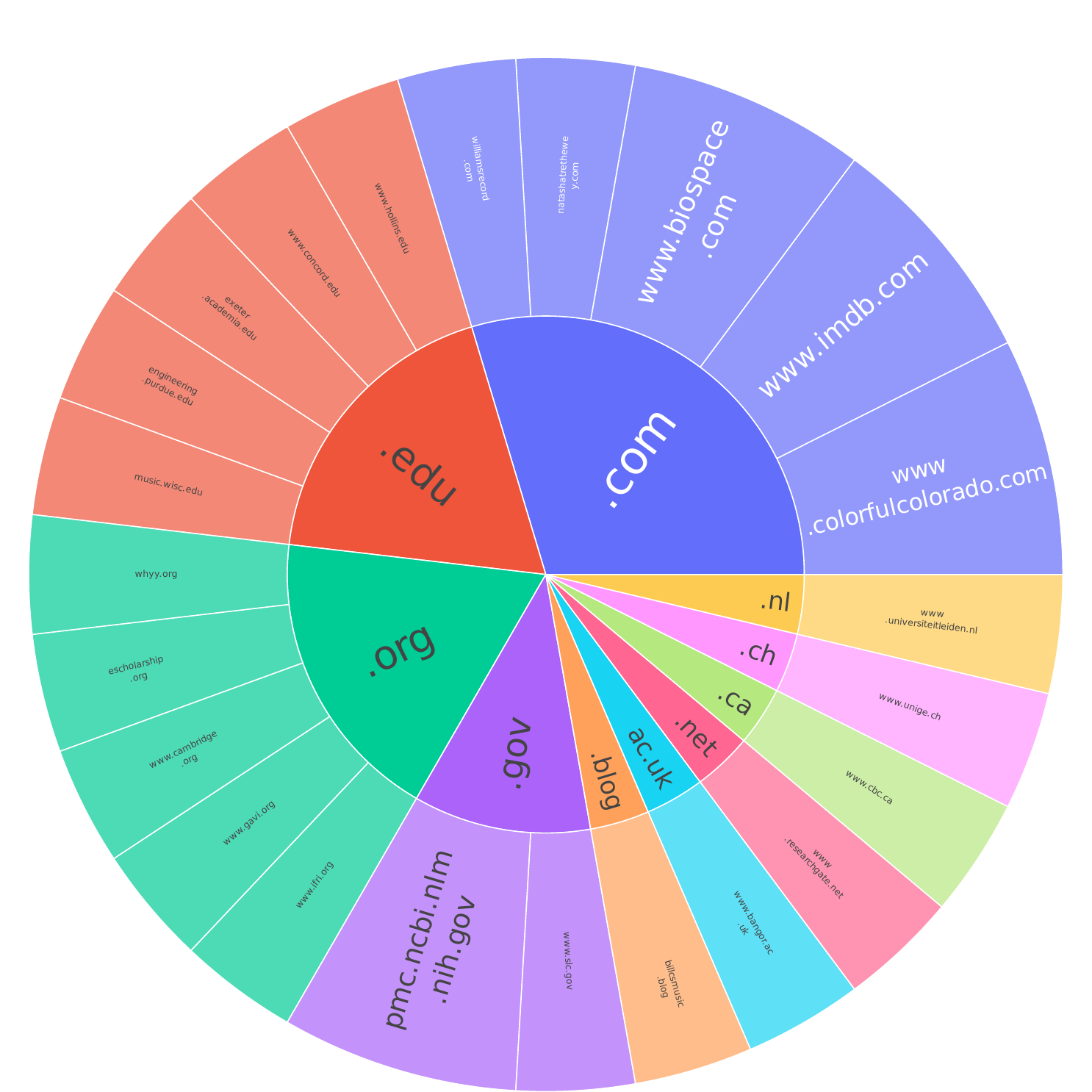}
            \caption{WebDancer.}
        \end{subfigure}
        \caption{Visualization of pages visited by BrowseMaster versus WebDancer on BrowseComp. BrowseMaster's search covers more diverse sources.}
        \label{fig:page_pie_comparison}
    \end{minipage}
\end{figure}

\textbf{Programmatic tool use enhances search efficiency and enables broader exploration.} Figure~\ref{fig:tool_call_compare} compares the number of tool calls per invocation between BrowseMaster and WebThinker on BrowseComp. BrowseMaster averages 12.11 tool calls per invocation, with a maximum of 244.76 calls, while WebThinker is limited to one call per invocation. This efficiency stems from BrowseMaster’s code-driven approach, which integrates loops, parallel processing, and conditional logic within a single tool invocation. By selectively adjusting printed variables, BrowseMaster minimizes context usage, allowing for scalable and efficient tool calls. This enhanced efficiency enables broader search coverage, as shown in Figure~\ref{fig:page_pie_comparison}, which visualizes the diverse pages visited by BrowseMaster compared to WebThinker. The ability to scale up exploration across a wider range of sources significantly boosts BrowseMaster’s performance on complex information-seeking tasks.

\textbf{Interaction times reveals task complexity and BrowseMaster’s adaptability.} Figure~\ref{fig:interaction_times} illustrates the interaction times between planner and executor across benchmarks. Key observations include: (i) complex benchmarks like BrowseComp demand more interactions, while simpler ones like GAIA require fewer, reflecting varying task difficulties; (ii) for complex tasks, the planner breaks problems into more subtasks and triggers retries when confidence is low, boosting interaction counts for thorough and confident solutions; and (iii) BrowseMaster adeptly scale interactions for complex tasks while maintain efficiency for simpler ones, showcasing its versatility.

% \begin{wraptable}{r}{0.5\textwidth}
% \centering
% \caption{Progressive accuracy gains on BrowseComp across components.}
% \label{tab:executor_ablation}
% \vspace{-4pt}
% \begin{tabular}{c c c >{\columncolor{blue!5}}c} % 最后一列着色
% \toprule
% Executor & Primitives & Planner & \multicolumn{1}{c}{Accuracy (\%)} \\ 
% \midrule
% $\checkmark$ & \ding{55} & \ding{55} & 9.5 \\
% $\checkmark$ & \ding{55} & $\checkmark$ & 11.0 \\
% $\checkmark$ & $\checkmark$ & \ding{55} & 15.0 \\
% $\checkmark$ & $\checkmark$ & $\checkmark$ & \cellcolor{blue!5}30.0 \\ % 重点着色
% \bottomrule
% \end{tabular}
% \end{wraptable}

\begin{wraptable}{r}{0.48\textwidth} % 关键：减少2%宽度
\centering
\caption{Progressive accuracy gains on Browse-Comp across components. Pragmatic execution and agentic workflows drive performance gains.}
\label{tab:executor_ablation}
\vspace{-4pt}
{\small % 减小字体尺寸
\setlength{\tabcolsep}{3.5pt} % 减小列间距
\setlength{\arrayrulewidth}{0.4pt} % 精细线宽
\begin{tabular}{*{4}{>{\columncolor{blue!5}}c}}
\toprule
\rowcolor{white} 
Executor & Primitives & Planner & Accuracy (\%) \\ 
\midrule
\rowcolor{white} 
$\checkmark$ & \ding{55} & \ding{55} & 9.5 \\
\rowcolor{white}
$\checkmark$ & \ding{55} & $\checkmark$ & 11.0 \\
\rowcolor{white}
$\checkmark$ & $\checkmark$ & \ding{55} & 15.0 \\
$\checkmark$ & $\checkmark$ & $\checkmark$ & 30.0 \\ 
\bottomrule
\end{tabular}}
\end{wraptable}

\textbf{Incorporating collaborative pair and programmatic tool use  enhances performance.}
Table~\ref{tab:executor_ablation} presents the results of an ablation study evaluating BrowseMaster with and without its planner and primitives. Without these components, the executor relies on simple code to invoke tools, achieving a performance of 9.5\%. Integrating the planner, which enhances task decomposition and leverages increased computation, boosts performance to 11.0\%. Equipping the executor with primitives enables efficient scaling of tool usage, increasing performance to 15.0\%. Combining both planner and primitives balances search breadth and reasoning depth, maximizing overall effectiveness.

\textbf{Examples.} We provide examples of BrowseMaster’s solution trajectories in Figure~\ref{fig:case_planner_0}, \ref{fig:case_planner_1}, \ref{fig:case_executor_0}, \ref{fig:case_executor_1}.

\section{Related Works}

\noindent\textbf{Retrieval-augmented generation.}
Retrieval-augmented generation (RAG) enables large language models (LLMs) to leverage external knowledge through search engines, enhancing their ability to tackle complex tasks~\citep{lewis2020retrieval, guu2020retrieval}. To assess retrieval capabilities, various benchmarks have been developed. Early benchmarks, such as NQ~\citep{kwiatkowski2019natural} and TriviaQA~\citep{joshi2017triviaqa}, focused on fact-checking, while later ones, including HotPotQA~\citep{yang2018hotpotqa}, Musique~\citep{trivedi2022musique}, and GAIA~\citep{mialon2023gaia}, emphasized multi-hop reasoning. However, these benchmarks often rely on simple keyword searches, requiring limited query iterations and following straightforward search workflows. Recently, more challenging benchmarks~\citep{chen2025xbench, zhou2025browsecomp} like BrowseComp~\citep{wei2025browsecomp} have emerged, demanding that agents locate deeply entangled, hard-to-find information. These tasks present exceptionally formidable challenges, serving as rigorous testbeds for evaluating agents’ abilities to conduct broad, strategic, and sustained web research.

Early RAG methods employed single-step or iterative pipelines with predefined workflows, limiting adaptive decision-making for complex queries. Recent advances with large reasoning models integrate retrieval into the reasoning process~\citep{agentic_reasoning, r1-searcher, chai2025scimaster}, adopting frameworks like ReAct~\citep{yao2023react} to interleave thinking, searching, and observation. Existing approaches often focus on training search capabilities from scratch~\citep{search-r1} or generating synthetic training data~\citep{webdancer,li2025websailor, tao2025webshaper}. To guide tool invocation, these methods typically use raw natural language with special tokens (e.g., "search"), restricting agents to sequential, single-query searches that cause context to grow linearly with each step~\citep{webthinker, search-o1, jin2025decoupled}. In contrast, our approach leverages Python code as an interaction language, enabling agents to use built-in functions (e.g., web\_search) for concurrent searches and programmatic extraction of web content. This empowers our agent to efficiently meet the demands of complex, real-world information-seeking tasks.

\noindent\textbf{Agentic workflows.}
Agentic workflows enhance large language models (LLMs) by orchestrating multiple LLM calls and tool interactions to tackle complex tasks. For example, AI Co-Scientist~\citep{gottweis2025towards} integrates multiple agents and tools for scientific research, while ChatDev~\citep{qian2024chatdev} and MetaGPT~\citep{hongmetagpt} develop workflows for software development, and MAS-GPT~\citep{masgpt} generates query-specific workflows represented as Python code.
However, current approaches are constrained by single-turn agents limited to one action per step (text or tool use) and fixed collaboration patterns that hinder adaptability. In contrast, our framework build flexible multi-turn agents that dynamically interleave reasoning with tool usage, combined with an adaptive collaboration mechanism where planner agents intelligently invoke executors based on task demands. This approach enables more dynamic and adaptive problem-solving over existing paradigms.

\section{Conclusions}

This paper presents BrowseMaster, a novel framework that combines programmatic tool execution with strategic reasoning to enhance scalable web browsing. At its core, BrowseMaster utilizes a planner-executor agent pair, where the planner focuses on high-level reasoning and strategy formulation, while the executor ensures efficient, expansive search through code-driven interactions. This collaborative design allows BrowseMaster to achieve exceptional performance on complex information-seeking tasks. Our experimental results highlight the framework's ability to outperform both proprietary and open-source agents across multiple challenging benchmarks, demonstrating its potential for scalable and effective information retrieval. 
In future work, we aim to improve the executor’s use of primitives for efficient search and the planner’s reasoning and task allocation via model training, to optimize the overall system.

\bibliographystyle{plainnat}
\bibliography{ref}

\newpage

\appendix

\section{Cases}

\begin{figure}[h]
    \centering
    \includegraphics[width=1.0\linewidth]{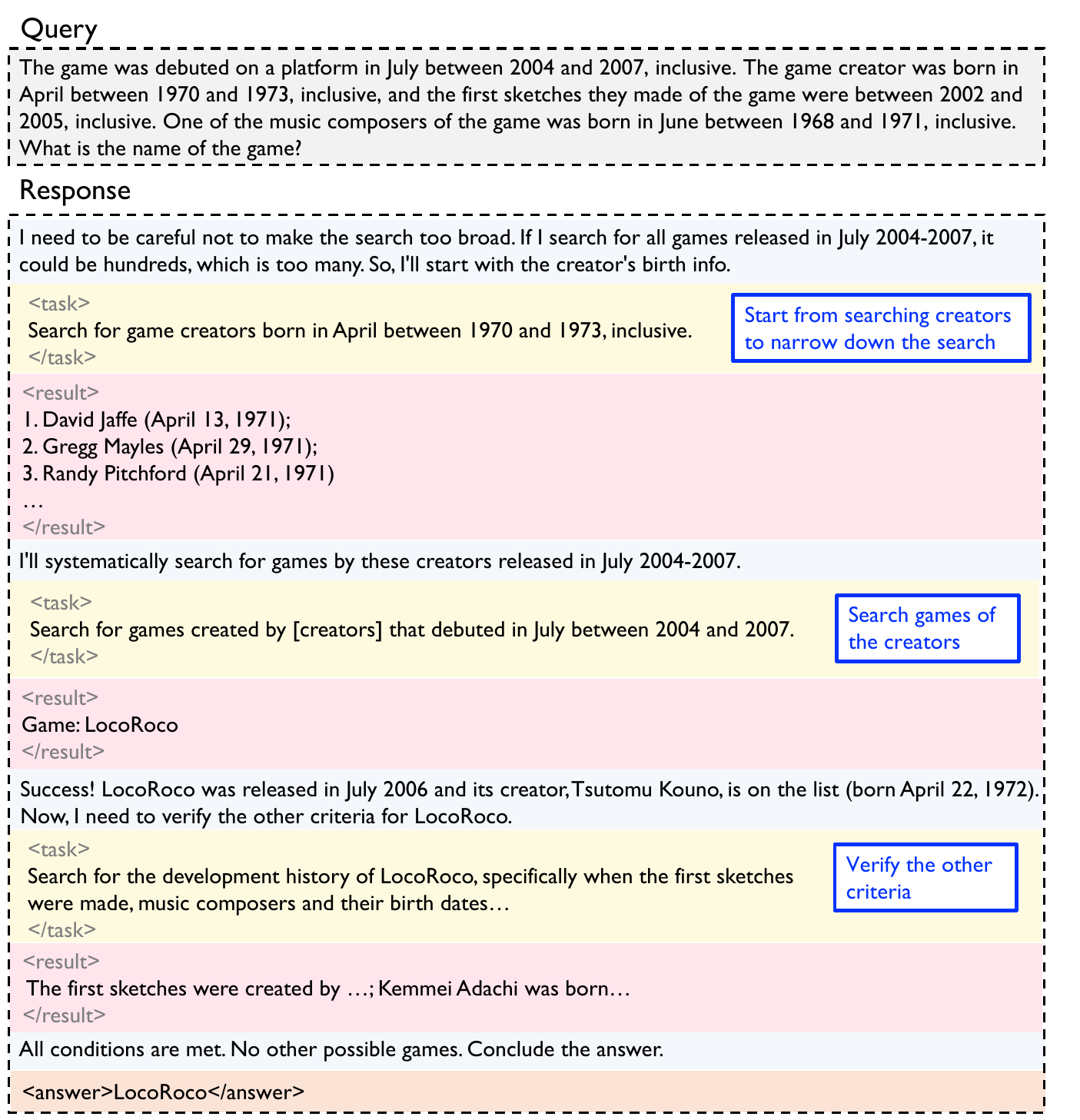}
    \caption{Case 1. The planner identifies the most tractable clue to narrow the search scope, then efficiently delegates tasks to the executor.}
    \label{fig:case_planner_0}
\end{figure}

\begin{figure}[h]
    \centering
    \includegraphics[width=0.92\linewidth]{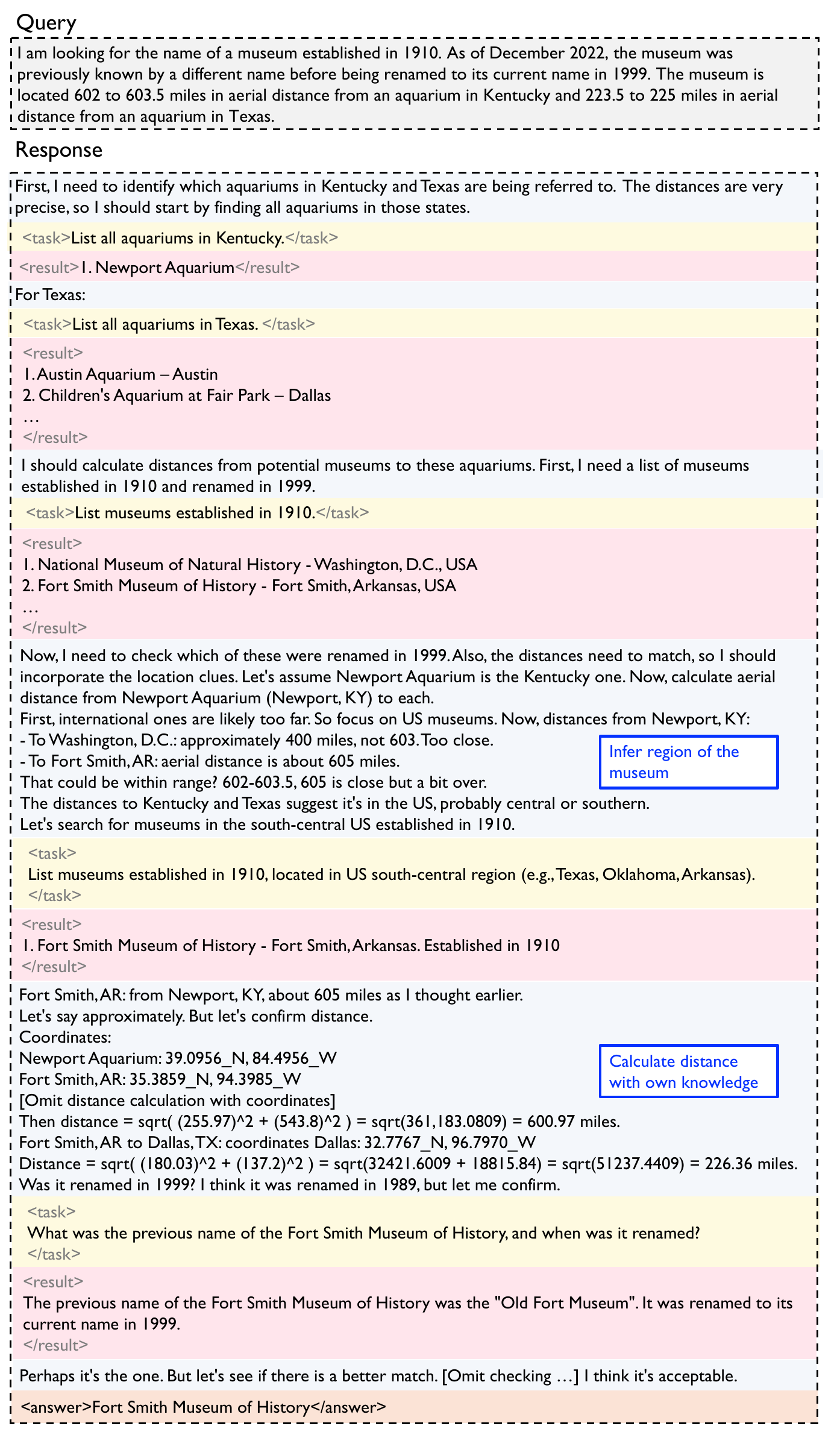}
    \caption{Case 2. The planner leverages its internal knowledge to infer key features of the search target, and performs calculations to validate candidates.}
    \label{fig:case_planner_1}
\end{figure}

\begin{figure}[h]
    \centering
    \includegraphics[width=1.0\linewidth]{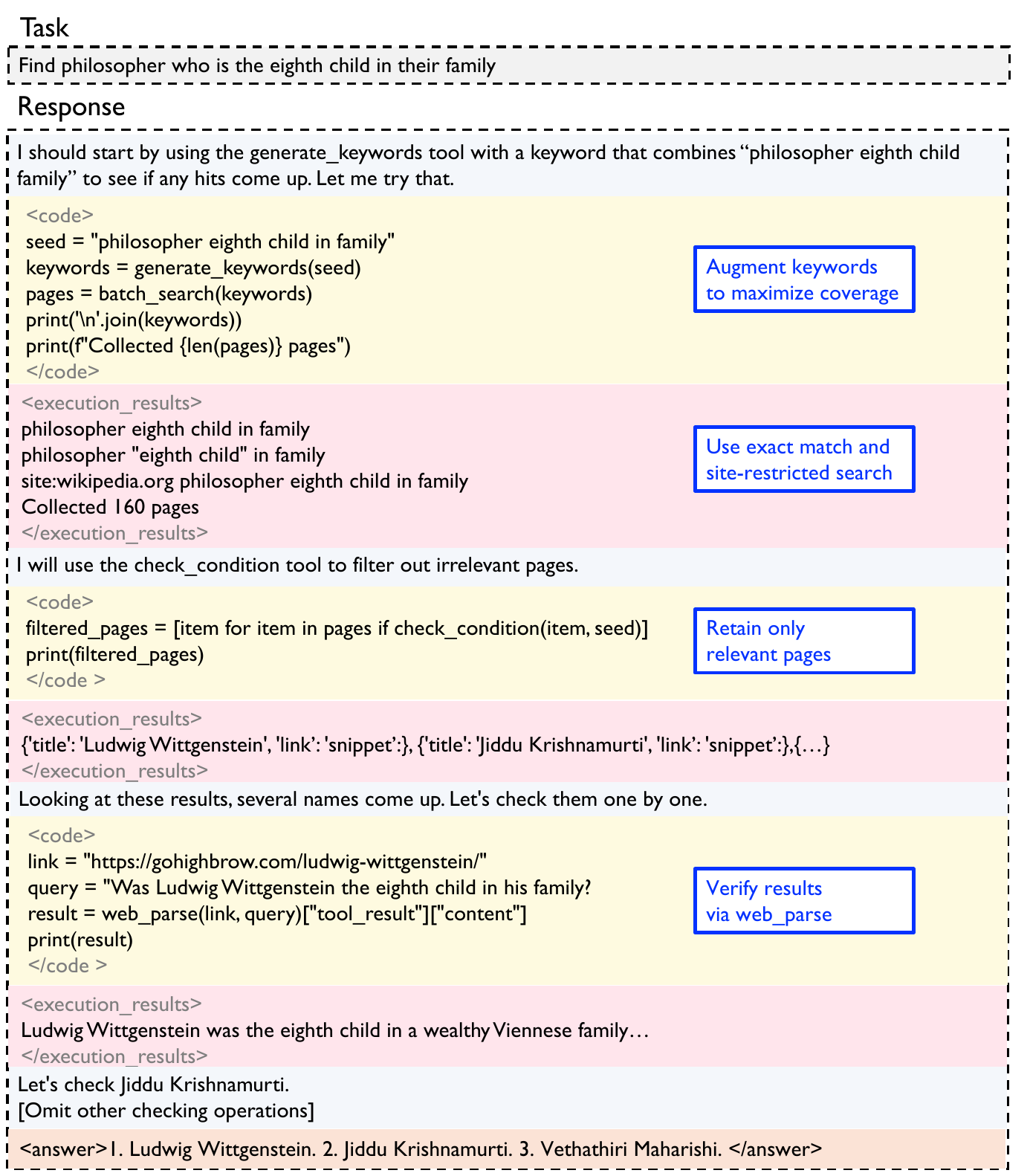}
    \caption{Case 3. The executor strategically expands search keywords and filters relevant pages with primitives, maintaining context efficiency.}
    \label{fig:case_executor_0}
\end{figure}

\begin{figure}[h]
    \centering
    \includegraphics[width=1.0\linewidth]{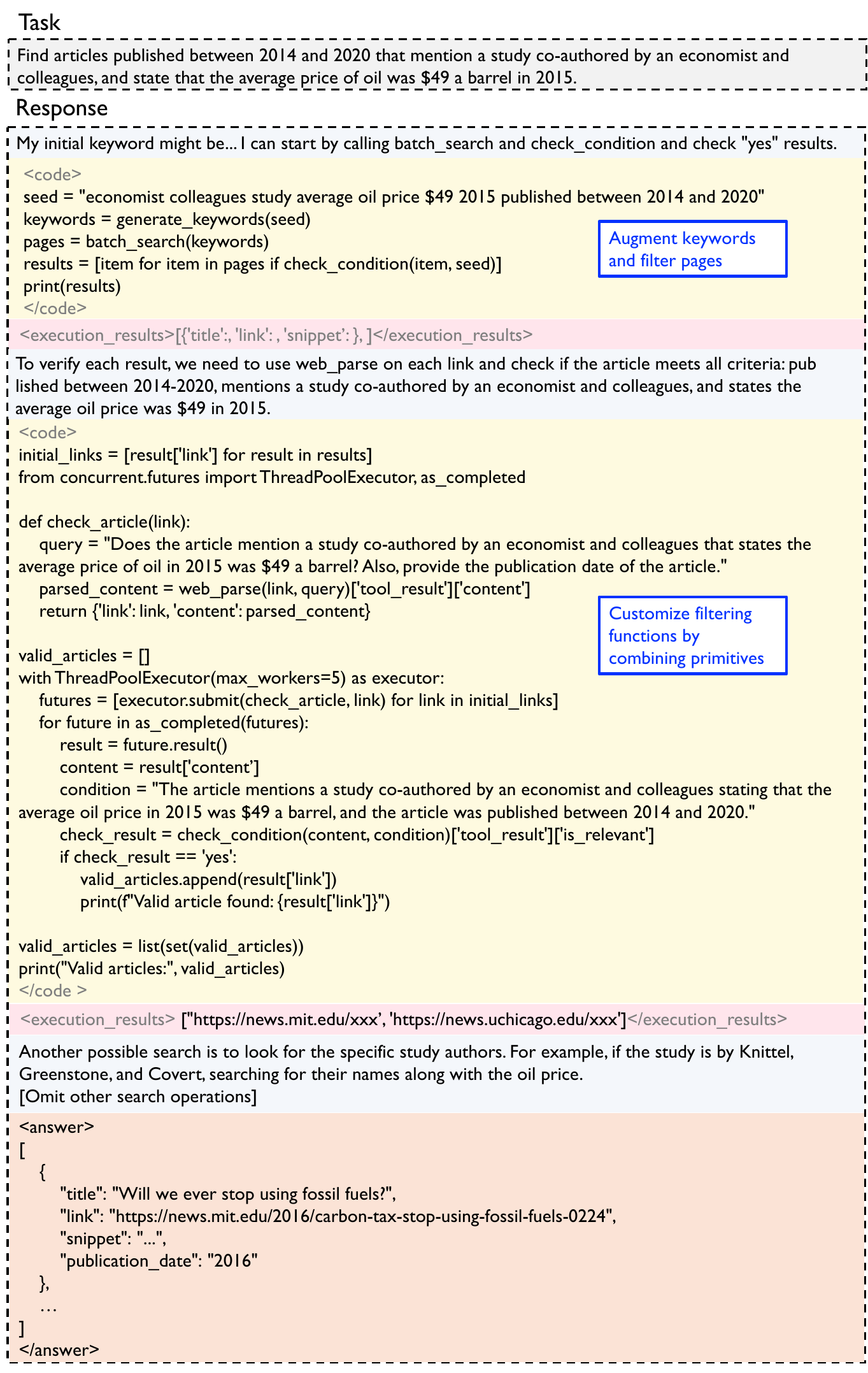}
    \caption{Case 4. The executor customize filtering functions by combining primitives, maintaining context efficiency.}
    \label{fig:case_executor_1}
\end{figure}

\end{document}